\documentclass{article}

\usepackage{microtype}
\usepackage{graphicx}
\usepackage{booktabs} 

\usepackage{colortbl}
\usepackage{xcolor}
\usepackage{xspace}

\usepackage{siunitx}
\usepackage{multirow}
\usepackage{float}
\usepackage{hyperref}

\usepackage[accepted]{icml2025}

\usepackage{amsmath}
\usepackage{amssymb}
\usepackage{mathtools}
\usepackage{amsthm}

\usepackage[capitalize,noabbrev]{cleveref}

\theoremstyle{plain}

\theoremstyle{definition}

\theoremstyle{remark}

\usepackage[utf8]{inputenc}
\usepackage[T1]{fontenc}
\usepackage{datetime2}
\usepackage{multirow}  
\usepackage{colortbl}  
\usepackage{listings}
\usepackage[textsize=tiny]{todonotes}

\icmltitlerunning{GeoVision Labeler}

\newcommand{\correspondingauthornote}[1]{%
  \renewcommand{\thefootnote}{}%
  \footnotetext{\textsuperscript{*}Corresponding author: #1}%
  \renewcommand{\thefootnote}{\arabic{footnote}}%
}

\begin{document}

\twocolumn[
\icmltitle{GeoVision Labeler: Zero-Shot Geospatial Classification with Vision and Language Models}



\icmlsetsymbol{corr}{*}

\begin{icmlauthorlist}
\icmlauthor{Gilles Quentin Hacheme}{ai4g-nairobi,corr}
\icmlauthor{Girmaw Abebe Tadesse}{ai4g-nairobi}
\icmlauthor{Caleb Robinson}{ai4g-seattle}
\icmlauthor{Akram Zaytar}{ai4g-nairobi}
\icmlauthor{Rahul Dodhia}{ai4g-seattle}
\icmlauthor{Juan M. Lavista Ferres}{ai4g-seattle}
\end{icmlauthorlist}

\icmlaffiliation{ai4g-nairobi}{Microsoft AI for Good Research Lab, Nairobi, Kenya}
\icmlaffiliation{ai4g-seattle}{Microsoft AI for Good Research Lab, Seattle, WA USA}

\icmlcorrespondingauthor{Gilles Quentin Hacheme}{ghacheme@microsoft.com}

\icmlkeywords{Zero-shot learning, Satellite image classification, Vision models, Large Language Models}

\vskip 0.3in
]

\correspondingauthornote{Gilles Quentin Hacheme (ghacheme@microsoft.com)}



\begin{abstract}
Classifying geospatial imagery remains a major bottleneck for applications such as disaster response and land-use monitoring--particularly in regions where annotated data is scarce or unavailable. Existing tools (e.g., RS-CLIP) that claim zero-shot classification capabilities for satellite imagery nonetheless rely on task-specific pretraining and adaptation to reach competitive performance. We introduce GeoVision Labeler (GVL), a strictly zero-shot classification framework: a vision Large Language Model (vLLM) generates rich, human-readable image descriptions, which are then mapped to user-defined classes by a conventional Large Language Model (LLM). This modular, and interpretable pipeline enables flexible image classification for a large range of use cases. We evaluated GVL across three benchmarks--SpaceNet v7, UC Merced, and RESISC45. It achieves up to 93.2\% zero-shot accuracy on the binary \textit{Buildings} vs. \textit{No Buildings} task on SpaceNet v7. For complex multi-class classification tasks (UC Merced, RESISC45), we implemented a recursive LLM-driven clustering to form meta-classes at successive depths, followed by hierarchical classification--first resolving coarse groups, then finer distinctions--to deliver competitive zero-shot performance. GVL is open-sourced at \url{https://github.com/microsoft/geo-vision-labeler} to catalyze adoption in real-world geospatial workflows.
\end{abstract}

\section{Introduction}
Geospatial image classification plays a pivotal role in remote sensing, supporting applications such as land use monitoring, urban planning, and disaster response~\cite{yin2021integrating, al2024integrating, navin2020comprehensive}. Traditional supervised learning methods rely on large, labeled datasets, which are often resource-intensive to compile or entirely unavailable, especially in scenarios such as disaster response~\cite{Mehmood2022, Li2022, Song2019, Li31122024}. Zero-shot learning presents a compelling alternative, allowing models to classify images without prior exposure to labeled examples of target classes~\cite{huadongreview2024, sun2024zero, mirza2023lafter}. Moreover, zero-shot classification can greatly accelerate human annotation workflows by providing timely weak labels that guide and streamline manual verification~\cite{wang2025pre, li2024leverage}.
While some models offer zero-shot capabilities, they often rely on pretraining with remote sensing data or synthetic pseudo-labels. 
Indeed, frameworks such as RS-CLIP~\cite{li2023rs} and SuperCLIP~\cite{damallasuperclip2025}, though designed for zero-shot remote sensing scene classification, adopt pretraining strategies thereby deviating from a strict zero-shot learning paradigm. In contrast, GeoVision Labeler (GVL) follows a strict zero-shot approach, requiring no fine-tuning or domain-specific customization.

GVL achieves this goal by combining vision Large Language Models (vLLMs) and Large Language models (LLMs) to perform zero-shot classification of geospatial images. GVL leverages the complementary strengths of vLLMs and LLMs to overcome the known limitation that vLLMs excel at generating rich, high-level descriptions but often struggle to assign inputs to a fixed taxonomy of labels~\cite{snaebjarnarson2025taxonomy, zhang2024visually, menon2022visual}. Given a geospatial image and user-specified categories, first, the vLLM generates a detailed textual description. The LLM then classifies the description into one of the predefined categories. This modular architecture allows the tool to adapt seamlessly to various classification tasks by simply updating the list of target classes, making it highly versatile for diverse remote sensing applications.
Additionally, because the classification is based on a human-readable description, users can readily understand the classification decisions, making GVL an inherently interpretable tool~\cite{ijcai2024p1025, menon2022visual, bilal2025llms}. 

Classification tasks in geospatial context that involve a large set of semantically related labels can be challenging, as models often confuse subtly different categories~\cite{sun2020fine, ye2021disentangling, he2017fine, ge2015subset}. To this end, We introduced a zero-shot LLM-based semantic clustering pipeline that optionally organizes labels into a handful of coarse \emph{meta-classes}, then refines within each group. By recursively applying (1) meta-class name generation, (2) meta-class assignment to each class, and (3) hierarchical extension, our approach reduces ambiguity among similar classes and achieves robust classification Overall Accuracy (OA) on more dissimilar meta-classes without any task-specific training.  

This paper evaluates GVL on three widely recognized remote sensing datasets: Spacenet v7~\citep{van2021spacenet}, UC Merced~\citep{neumann2019domain}, and RESISC45~\citep{yang2010bag, neumann2019domain}. Through zero-shot experiments, we assessed GVL's performance across different model configurations and compared with UC Merced and RESISC45 results reported in recent studies~\cite{corley2024revisiting, li2023rs}. Our findings demonstrate GVL's strong potential to achieve competitive classification OA, particularly in scenarios where label classes are well-separated and exhibit minimal overlap. For example, on the binary \textit{Buildings} vs. \textit{No Buildings} task from SpaceNet v7, our pipeline achieves up to 93.2\% zero-shot OA.  Moreover, by merging semantically similar classes via our clustering method, GVL attains up to 86.4\% and 84.3\% OA respectively on meta-classes from UC Merced and RESISC45. 

GVL offers a fast, interpretable, and adaptable zero-shot classification framework for real-world satellite image analysis. It can be used for rapid tagging and visual search of satellite imagery--crucial when labels are scarce or time-to-action is critical. The source code for GVL is openly accessible from Github\footnote{\url{https://github.com/microsoft/geo-vision-labeler}}, facilitating further research and application.

\section{Background}
The classification of geospatial images has advanced significantly with the advent of deep learning, yet the reliance on labeled data remains a bottleneck~\cite{tao2020remote, li2020rs}. 

Recent research has explored the application of pretrained vision models for zero-shot applications in remote sensing~\cite{li2023rs, damallasuperclip2025}. For instance, RS-CLIP achieves impressive accuracies of 95.94\% on UC Merced and 85.76\% on RESISC45~\cite{li2023rs}. However, its flexibility remains limited, partly due to the demand for an additional image–text pretraining stage, the generation of pseudo-labels from unlabeled imagery, and a multi-stage curriculum learning step to progressively adapt the model. SuperCLIP similarly departs from a pure zero-shot approach: it enhances feature learning with super-resolution modules and semantic attribute-guided transformers, relying on domain-specific adaptations and additional training~\cite{damallasuperclip2025}. These approaches, while effective, rely on subsequent pretraining.

Our work provides a strict zero-shot learning framework by integrating a vLLM and an LLM classifier into a single, end-to-end pipeline that generates detailed image descriptions and directly maps them to user-defined classes. Unlike approaches that depend on additional adaptation, GVL does not require a task-specific training or a subsequent tuning--making it a broader, more efficient solution for rapid deployment across diverse geospatial classification tasks.

Existing vLLMs such as Kosmos-2 and Llama 3.2 (Vision-Instruct) are pre-trained on vast amount of images and text data, enabling them to understand and generate relationships between visual and textual information~\cite{peng2023kosmos, grattafiori2024llama}. As a result, these models have the potential to excel in zero-shot image understanding tasks. On the other hand, LLMs, including GPT-4o and Llama models, are proficient in natural language processing, capable of classifying textual descriptions into predefined categories based on semantic reasoning~\cite{achiam2023gpt, grattafiori2024llama}.
GVL leverages the complementary capabilities of vLLMs and LLMs: while vLLMs excel at generating rich image descriptions, they often struggle with mapping inputs to a fixed label taxonomy--a task at which LLMs demonstrate superior capability~\cite{snaebjarnarson2025taxonomy, zhang2024visually, menon2022visual}.
Finally, because GVL bases its classifications on human-readable descriptions, users can easily interpret the model's decisions--making it an inherently transparent and interpretable tool~\cite{ijcai2024p1025, menon2022visual, bilal2025llms}.

\section{Datasets}
We benchmarked GVL across three remote sensing datasets described below.

\textbf{SpaceNet v7:}
We evaluated GVL on $59$ ($1024\times1024$) scenes from SpaceNet V7, a multi-temporal building footprint dataset accross 100 locations~\cite{van2021spacenet}. We split each image into 9 patches resulting in $59 \times 9 = 531$ image patches, each labeled as \textit{Buildings} if any pixel overlapped with a building footprint, or \textit{No Buildings} otherwise. A random timestamp was selected per scene to account for the dataset's temporal nature. 

\textbf{UC Merced:}
It consists of $2,100$ RGB images, each includes $256 \times 256$ pixels, across $21$ land use classes, including agricultural, forest, and residential areas~\citep{neumann2019domain}. We used the $420$ images provided in the test set in our experiments. 

\textbf{RESISC45:}
It is a benchmark for remote sensing image scene classification containing $31,500$ RGB images of $256 \times 256$ pixels, spanning $45$ scene classes such as airports, forests, and residential areas, with $700$ images per class~\citep{yang2010bag, neumann2019domain}. We used the $6,300$ image scenes provided in the test set in our experiments. 


\section{Methodology}

\paragraph{Description + Classification Pipeline:}
\begin{figure*}[htbp]
    \centering
    \includegraphics[width=0.75\linewidth]{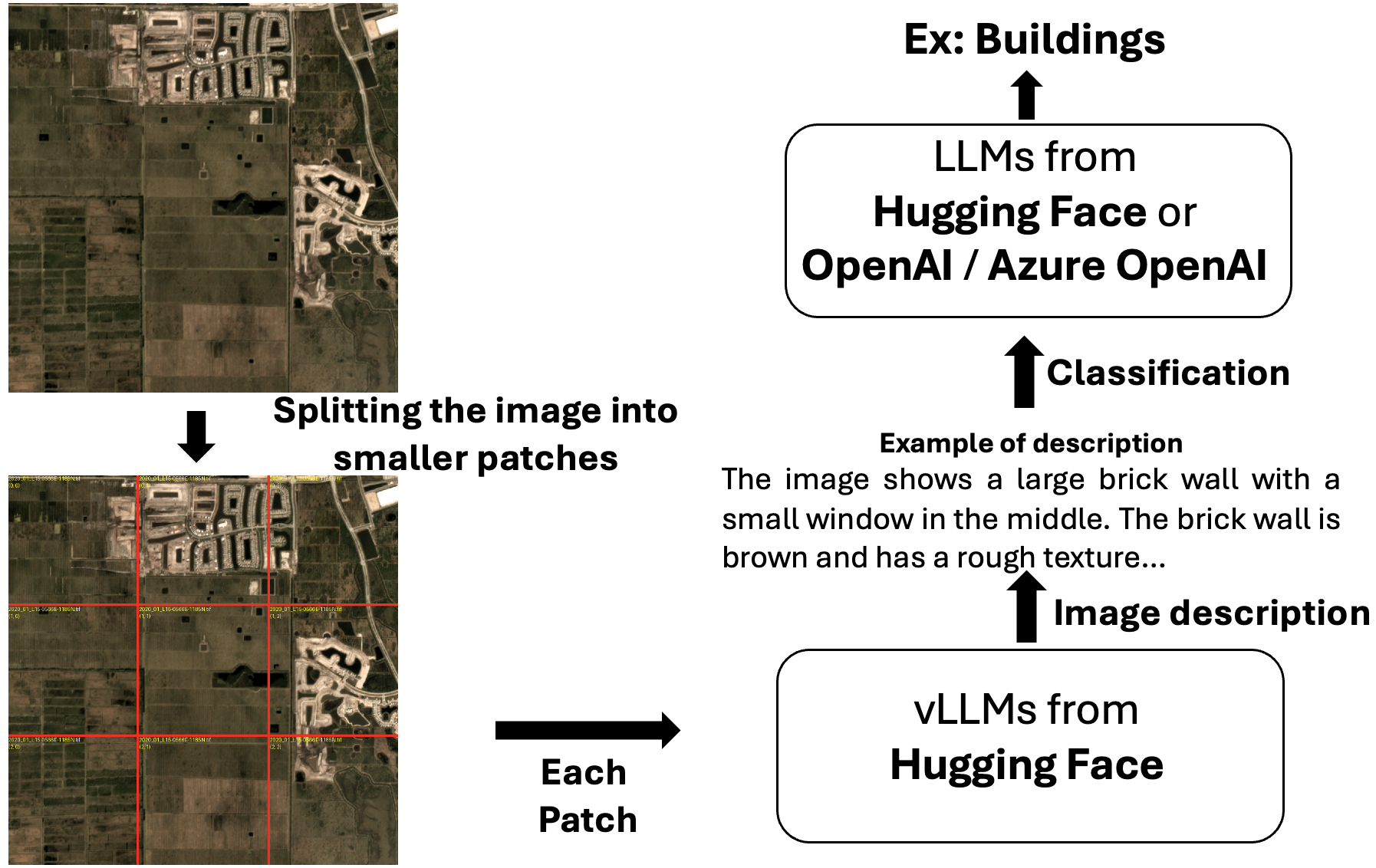}
    \caption{GeoVision Labeler (GVL) pipeline. GVL takes any image as input and generates a classification label from a set of user-provided classes. It uses a vLLM from Hugging Face (e.g., \textit{microsoft/kosmos-2-patch14-224}) to generate a detailed description of the image, and then uses an LLM to classify. The CLIP model is used as a fallback when the primary pipeline (vLLM + Classifier) fails to generate a valid label (i.e., a label not included in the list of classes). The CLIP model classifies the image directly by calculating cosine similarity between its embedding and text embeddings derived from class label prompts.}
    \label{fig:pipeline}
\end{figure*}

GVL employs a two-stage pipeline, \textit{Image description} and \textit{Classification}, to achieve zero-shot classification of geospatial images, leveraging the strengths of vLLMs and LLMs (see Figure \ref{fig:pipeline}).
The classification process begins with image description generation (stage 1). Optionally, images can be patched into smaller grid cells, each of which is used individually to obtain localized descriptions by prompting the vLLM. A vLLM, such as Kosmos-2, processes the prompt and the input image patch to produce a detailed textual description. The prompt provided to the vLLM typically includes a context statement, such as \textit{``This is a satellite image,''} followed by a directive to produce a detailed description, ensuring the output captures relevant visual features. The prompt may also include the filename, if it contains embedded geographic information (e.g., \textit{L15-2044E-0928N}), and the list of classes.
The generated description is then passed to an LLM, such as GPT-4o, for classification (stage 2). The LLM is prompted with the description and a list of user-defined classes, tasked with selecting the most appropriate category. If the LLM produces an invalid label (i.e., not in the provided class list), the CLIP model~\cite{radford2021learning} serves as a fallback, directly classifying the image by computing cosine similarity scores between the image and text prompts constructed from the class labels.

\begin{figure*}[htbp]
    \centering
    \includegraphics[width=0.75\linewidth]{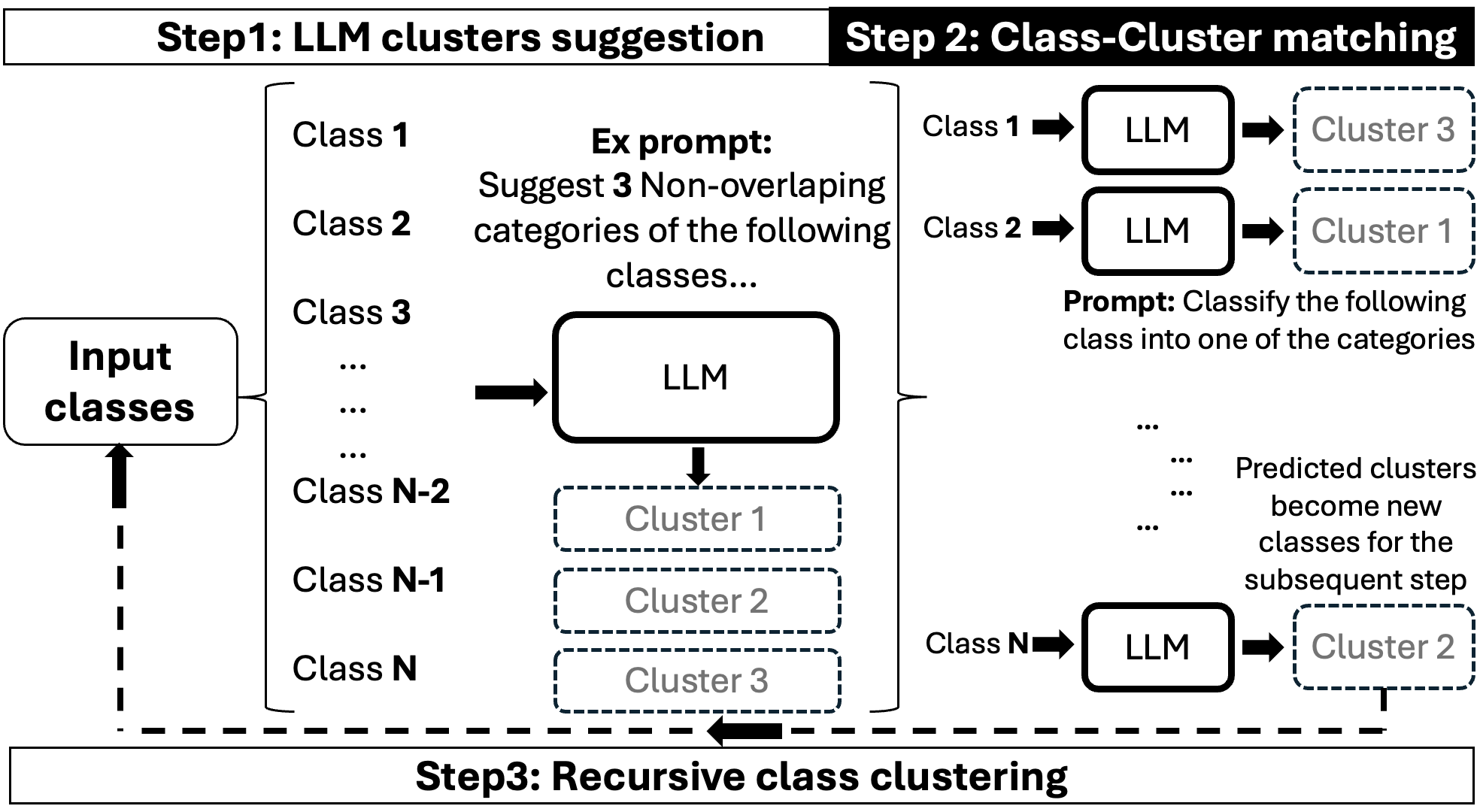}
    \caption{Overview of the recursive class clustering pipeline using an LLM. Step 1: The LLM is prompted to suggest semantically coherent cluster names from a list of input classes. Step 2: Each class is individually matched to one of the suggested clusters using an LLM-based classification. Step 3: The process is recursively applied to each cluster to build a hierarchical taxonomy of classes.}
    \label{fig:class_clustering}
\end{figure*}

\paragraph{Semantic Hierarchical Clustering of Classes:}

When faced with a large set of semantically related labels (e.g., 21 UC Merced land-use categories), it can be advantageous to first group them into a small number of \emph{meta-classes} before performing classification. We propose a zero-shot, LLM-based semantic clustering pipeline in 3 steps. (See Figure \ref{fig:class_clustering})

\textit{Step 1: Generating Meta-classes Names}

Let $\mathcal{C} = \{c_1, c_2, \dots, c_N\}$
be the set of original class labels ($N$), and let $K$ be the desired number of meta-classes.  
In the first step, the LLM is prompted to generate $K$ meta-class names by issuing the following request:
\begin{quote}\small
\emph{Suggest $K$ non-overlapping category names for the following labels based on semantic similarity.  Output in the form \texttt{Cluster\_1: [Name]}, \dots, \texttt{Cluster\_K: [Name]}.}
\end{quote}
The $K$ resulting names $\{M_j\}_{j=1}^K$ are then parsed and lightly cleaned (e.g., punctuation is removed and extraneous whitespace is trimmed).  

\textit{Step 2: Matching classes to meta-classes (class clusters)}

In the second step, each original label $c_i \in \mathcal{C}$ is individually matched to a meta-class $M_j$ by prompting the model to:

\begin{quote}\small
\emph{Assign this label to one of the categories $\{M_1,\dots,M_K\}$.  Output in the form \texttt{Cluster: [Name]}.}
\end{quote}

If the returned category exactly matches one of the meta-classes, it is assigned accordingly; if not, a simple substring-matching heuristic is applied, and any remaining ambiguous cases are placed into an \textit{Unknown} bucket.  Finally, any empty meta-classes are discarded to yield the final set of clusters.
This produces a mapping $\mathcal{C}\;\longrightarrow\;\{M_1,\dots,M_{\tilde K}\},$
with $\tilde K \le K$.

\textit{Step 3: Recursive Extension for Hierarchy}

To obtain a hierarchy of depth $D$, we apply the above two-step procedure recursively.  Given a sequence of cluster sizes $[K_1, K_2, \dots, K_D]$:
\begin{itemize}
  \item At level 1, split $\mathcal{C}$ into $K_1$ meta-classes $\{M_j^{(1)}\}$.
  \item For each $M_j^{(1)}$, split into $K_2$ sub-classes $\{M_{j,k}^{(2)}\}$.
  \item Continue until depth $D$ is reached or a subset is empty.
\end{itemize}

\begin{figure*}[htbp]
    \centering
    \includegraphics[width=0.75\linewidth]{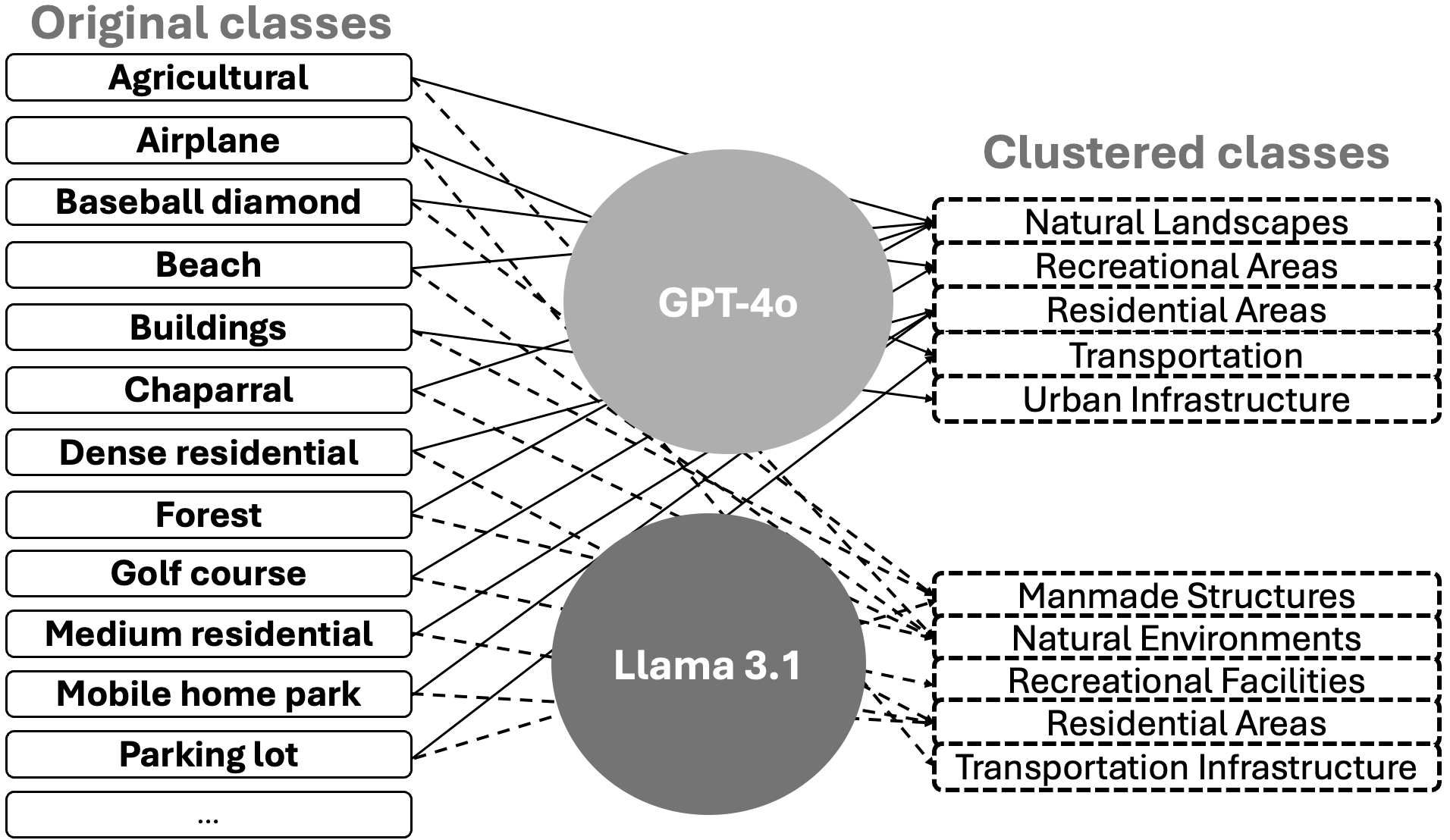}
    \caption{This diagram visualizes the semantic clustering of 12 out of the 21 original land use and land cover classes from the UC Merced dataset using two large language models: GPT-4o and Llama 3.1. On the left, the original dataset classes--such as Agricultural, Airplane, Beach, Forest, and Parking lot--are shown. These are mapped into higher-level semantic clusters, shown on the right.}
    \label{fig:ucm_class_clustering}
\end{figure*}

We applied this class clustering method to both the UC Merced and RESISC45 datasets, using $D=1$ and $D=2$, respectively.
An example of this clustering strategy performed on the 21 classes of the UC Merced dataset is shown in Figure \ref{fig:ucm_class_clustering}. The complete lists of meta-classes for both datasets are provided in Tables \ref{tab:ucmerced_meta_classes} and \ref{tab:resisc45_meta_classes} in the Appendix.

\paragraph*{Implementation and Evaluation Metric:}
GVL is implemented in Python, utilizing the Hugging Face (HF) Transformers library for open-source vLLM and LLM integration, and the OpenAI and Azure OpenAI APIs for accessing closed source models like GPT-4o. The tool's modular design allows for easy substitution of different vLLMs and LLMs, enabling experimentation with various model combinations.
For our experiments, we applied GVL to Spacenet v7, UC Merced, and RESISC45 in a zero-shot setting, without any additional fine-tuning. 
We experimented with two vLLMs including Kosmos 2\footnote{HF: \texttt{microsoft/kosmos-2-patch14-224}} and Llama 3.2\footnote{HF: \texttt{meta/Llama 3.2-11b-vision-instruct}} and three LLM classifiers including Llama 3.1\footnote{HF: \texttt{meta/llama-3.1-8b-instruct}}, Phi-3\footnote{HF: \texttt{phi-3-mini-4k-instruct}}, and GPT-4o~\cite{peng2023kosmos2, grattafiori2024llama, abdin2024phi, achiam2023gpt}.
We evaluated performance using classification Overall Accuracy (OA), with comparisons to existing benchmark results reported in recent literature for both UC Merced and RESISC45~\cite{corley2024revisiting, li2023rs}.

\section{Results}
\begin{table}[ht!]
\centering
\caption{Overall Accuracy (OA) for SpaceNet v7 satellite image patches from 59 scenes, each divided into 9 patches labeled as \textit{Buildings} or \textit{No Buildings} based on ground truth labels, with a random timestamp selected per scene. Best OA scores for each classifier and vision model from GeoVision Labeler (GVL) are highlighted in bold. The \textit{Classes} column indicates whether the list of classes is inserted into the vLLM's prompt. \textit{Other} refers to vision models that are used in a standalone way (i.e., not part of GVL).}
\resizebox{0.5\textwidth}{!}{%
\begin{tabular}{lcccccc}
\toprule
 &&&& \multicolumn{3}{c}{\textbf{Vision Model}} \\
\cmidrule(lr){5-7}
{\textbf{Pipeline}} & {\textbf{Classifier}} & {\textbf{Classes}} & {\textbf{Geo-context}} & \multicolumn{1}{c}{\textbf{Other}} & \multicolumn{1}{c}{\textbf{Kosmos 2}} & \multicolumn{1}{c}{\textbf{Llama 3.2}} \\
\midrule
\multirow{1}{*}{CLIP} & - & - & - & 0.588 & - & - \\
\cmidrule(lr){1-7}
\multirow{12}{*}{GVL (Ours)} & \multirow{4}{*}{Llama-3.1} & \multirow{2}{*}{\textcolor{black}{$\checkmark$}} & \textcolor{black}{$\times$} & - & 0.859 & 0.776 \\
& & & \textcolor{black}{$\checkmark$} & - & \textbf{0.910} & 0.699 \\
\cmidrule(lr){3-7}
& &  \multirow{2}{*}{\textcolor{black}{$\times$}} & \textcolor{black}{$\times$} & - & 0.889 & \textbf{0.821} \\
& & & \textcolor{black}{$\checkmark$} & - & 0.859 & 0.751 \\
\cmidrule(lr){2-7}
& \multirow{4}{*}{Phi-3} &  \multirow{2}{*}{\textcolor{black}{$\checkmark$}} & \textcolor{black}{$\times$} & - & \textbf{0.932} & 0.857 \\
& & & \textcolor{black}{$\checkmark$} & - & 0.928 & 0.902 \\
\cmidrule(lr){3-7}
& &  \multirow{2}{*}{\textcolor{black}{$\times$}} & \textcolor{black}{$\times$} & - & 0.928 & 0.912 \\
& & & \textcolor{black}{$\checkmark$} & - & \textbf{0.932} & \textbf{0.927} \\
\cmidrule(lr){2-7}
& \multirow{4}{*}{GPT-4o} &  \multirow{2}{*}{\textcolor{black}{$\checkmark$}} & \textcolor{black}{$\times$} & - & 0.878 & 0.789 \\
& & & \textcolor{black}{$\checkmark$} & - & \textbf{0.917} & 0.799 \\
\cmidrule(lr){3-7}
& &  \multirow{2}{*}{\textcolor{black}{$\times$}} & \textcolor{black}{$\times$} & - & 0.896 & \textbf{0.832} \\
& & & \textcolor{black}{$\checkmark$} & - & 0.876 & 0.783 \\
\bottomrule
\end{tabular}%
}
\label{tab:spacenet}
\end{table}

\begin{table}[ht!]
\centering
\caption{Overall Accuracy (OA) for UC Merced satellite image patches, evaluated in a zero-shot setting and compared with a fine-tuned ResNet50 with ImageNet weights baseline~\cite{corley2024revisiting}. Best OA scores for each classifier and vision model from GeoVision Labeler (GVL) are highlighted in bold. The \textit{Classes} column indicates whether the list of classes is inserted into the vLLM's prompt. \textit{Other} refers to vision models that are used in a standalone way (i.e., not part of GVL).}
\resizebox{0.5\textwidth}{!}{%
\begin{tabular}{lccccc}
\toprule
 &&& \multicolumn{3}{c}{\textbf{Vision Model}} \\
\cmidrule(lr){4-6}
{\textbf{Pipeline}} & {\textbf{Classifier}} & {\textbf{Classes}} & \multicolumn{1}{c}{\textbf{Other}} & \multicolumn{1}{c}{\textbf{Kosmos 2}} & \multicolumn{1}{c}{\textbf{Llama 3.2}} \\
\midrule
\multirow{1}{*}{CLIP} & - & - & 0.710 & - & - \\
\cmidrule(lr){1-6}
\multirow{1}{*}{ResNet50 (ImageNet)} & \multirow{2}{*}{-} & \multirow{2}{*}{-} & \multirow{2}{*}{0.907}  & \multirow{2}{*}{-} & \multirow{2}{*}{-} \\
~\cite{corley2024revisiting} & & & & &\\
\multirow{1}{*}{RS-CLIP~\cite{li2023rs}} & - & - & 0.959 & - & - \\
\cmidrule(lr){1-6}
\multirow{6}{*}{GVL (Ours)} & \multirow{2}{*}{Llama-3.1} & \textcolor{black}{$\checkmark$} & - & 0.514 & 0.600 \\
& & \textcolor{black}{$\times$} & - & \textbf{0.619} & \textbf{0.610} \\
\cmidrule(lr){2-6}
& \multirow{2}{*}{Phi-3} & \textcolor{black}{$\checkmark$} & - & 0.521 & \textbf{0.626} \\
& & \textcolor{black}{$\times$} & - & \textbf{0.648} & 0.624 \\
\cmidrule(lr){2-6}
& \multirow{2}{*}{GPT-4o} & \textcolor{black}{$\checkmark$} & - & 0.517 & \textbf{0.714} \\
& & \textcolor{black}{$\times$} & - & \textbf{0.710} & 0.710 \\

\bottomrule
\end{tabular}%
}
\label{tab:ucm}
\end{table}

We compared GVL's two-stage pipeline against the vanilla CLIP baseline and analyze the effects of prompt design, including class enumeration in the vLLM's prompt. Moreover, with SpaceNet v7, we explored the effect of geo-context injection in the vLLM's prompt. For the more complex multi-class datasets--UC Merced and RESISC45--we additionally explored a hierarchical classification strategy to reduce confusion among visually similar labels.

\textbf{SpaceNet v7:}  
Table~\ref{tab:spacenet} shows Overall Accuracy (OA) results for SpaceNet v7.  First, the vanilla CLIP baseline achieves only $0.588$ accuracy. GVL's two-stage pipeline yields higher accuracy across all model combinations. Inserting the two target labels (\textit{Buildings} vs. \textit{No Buildings}) to the vLLM prompt generally boosts accuracy, while further injecting geo-context from the image filename drives Kosmos 2's OA from $0.859$ to $0.910$ and lifts GPT-4o's OA from $0.878$ to $0.917$.
Kosmos 2 + Phi-3 with either classes (Classes=$\checkmark$, Geo-context=$\times$) or geo-context (Classes=$\times$, Geo-context=$\checkmark$) in the vLLM's prompt delivers the best zero-shot result ($0.932$). Kosmos 2 + GPT-4o follows at $0.917$, and Kosmos 2 + Llama-3.1 peaks only when both classes and geo-context (Classes=$\checkmark$, Geo-context=$\checkmark$) are provided. Across all settings, Kosmos 2 outperforms Llama 3.2 by up to $20$ percentage points, highlighting its stronger visual grounding.

\textbf{UC Merced:}
Table~\ref{tab:ucm} shows that CLIP achieves $0.710$ in OA--about $80$\% of a fine-tuned ResNet50 ($0.907$) and $74$\% of RS-CLIP ($0.959$)~\cite{corley2024revisiting, li2023rs}--showing that generic image–text similarity captures much of UC Merced's land-use signals. 
The best GVL OA score (0.714) is achieved by combining the Llama 3.2 vLLM with the GPT-4o classifier, using prompts that enumerate classes within the vLLM's prompt (Classes=$\checkmark$).
However, supplying all 21 classes in the vLLM's prompt often lowers accuracy: Kosmos 2 + Llama-3.1 drops by $\approx11$ percentage points ($0.619$ $\rightarrow$ $0.514$), Kosmos 2 + Phi-3 drops by $\approx13$ percentage points ($0.648$ $\rightarrow$ $0.521$), and Kosmos 2 + GPT-4o by $\approx19$ percentage points ($0.710 \rightarrow 0.517$) when classes are forced. This indicates that long class lists can dilute descriptive quality, especially for Kosmos 2, whereas Llama 3.2's performance remains roughly stable.
The confusion matrix for GVL with Llama 3.2 as vLLM and GPT-4o as classifier, including classes (Classes=$\checkmark$), (see Figure \ref{fig:cm-ucm-llama32-gpt4o-classes} in the Appendix) suggests that the lower performance of GVL on UC Merced is likely due to confusion among visually similar categories within the 21 classes.

\begin{table}[ht]
\centering
\caption{Overall Accuracy (OA) for hierarchical classification of UC Merced satellite images. The original 21 classes were clustered into 5 broader classes at depth $D=0$.  $D=1$ refers to classifying original classes. Best OA scores for each classifier and vision model are highlighted in bold. The \textit{Classes} column indicates whether the list of classes is inserted into the vLLM's prompt.}

\resizebox{0.5\textwidth}{!}{%
\begin{tabular}{l l c c  c c}
\toprule
 & & & & \multicolumn{2}{c}{\textbf{Vision Model}} \\
 \cmidrule(lr){5-6}
\textbf{Clustering} & \textbf{Classifier} & \textbf{Classes} & \textbf{Depth} & \textbf{Kosmos 2} & \textbf{Llama 3.2} \\
\midrule
\multirow{12}{*}{Llama-3.1}
  & \multirow{4}{*}{Llama-3.1}
    & \multirow{2}{*}{\textcolor{black}{$\times$}} & 0 & \textbf{0.738} & \textbf{0.755} \\
  &    &                          & 1 & 0.498 & 0.510 \\
  \cmidrule(lr){3-6}
  &    & \multirow{2}{*}{\textcolor{black}{$\checkmark$}} & 0 & 0.700 & 0.726 \\
  &    &                          & 1 & 0.452 & 0.460 \\
 \cmidrule(lr){2-6}
  & \multirow{4}{*}{Phi-3}
    & \multirow{2}{*}{\textcolor{black}{$\times$}} & 0 & \textbf{0.612} & \textbf{0.617} \\
  &    &                          & 1 & 0.393 & 0.391 \\
  \cmidrule(lr){3-6}
  &    & \multirow{2}{*}{\textcolor{black}{$\checkmark$}} & 0 & 0.562 & 0.614 \\
  &    &                          & 1 & 0.350 & 0.343 \\
 \cmidrule(lr){2-6}
  & \multirow{4}{*}{GPT-4o}
    & \multirow{2}{*}{\textcolor{black}{$\times$}} & 0 & \textbf{0.679} & 0.679 \\
  &    &                          & 1 & 0.483 & 0.507 \\
  \cmidrule(lr){3-6}
  &    & \multirow{2}{*}{\textcolor{black}{$\checkmark$}} & 0 & 0.643 & \textbf{0.681} \\
  &    &                          & 1 & 0.457 & 0.502 \\
\midrule
\multirow{12}{*}{GPT-4o}
  & \multirow{4}{*}{Llama-3.1}
    & \multirow{2}{*}{\textcolor{black}{$\times$}} & 0 & \textbf{0.795} & 0.743 \\
  &    &                          & 1 & 0.543 & 0.460 \\
  \cmidrule(lr){3-6}
  &    & \multirow{2}{*}{\textcolor{black}{$\checkmark$}} & 0 & 0.762 & \textbf{0.760} \\
  &    &                          & 1 & 0.548 & 0.460 \\
 \cmidrule(lr){2-6}
  & \multirow{4}{*}{Phi-3}
    & \multirow{2}{*}{\textcolor{black}{$\times$}} & 0 & \textbf{0.857} & \textbf{0.807} \\
  &    &                          & 1 & 0.595 & 0.543 \\
  \cmidrule(lr){3-6}
  &    & \multirow{2}{*}{\textcolor{black}{$\checkmark$}} & 0 & 0.771 & 0.795 \\
  &    &                          & 1 & 0.510 & 0.486 \\
 \cmidrule(lr){2-6}
  & \multirow{4}{*}{GPT-4o}
    & \multirow{2}{*}{\textcolor{black}{$\times$}} & 0 & \textbf{0.864} & 0.807 \\
  &    &                          & 1 & 0.662 & 0.602 \\
  \cmidrule(lr){3-6}
  &    & \multirow{2}{*}{\textcolor{black}{$\checkmark$}} & 0 & 0.757 & \textbf{0.836} \\
  &    &                          & 1 & 0.593 & 0.581 \\
\bottomrule
\end{tabular}%
}
\label{tab:hierarchical_ucm_grouped}
\end{table}

Table~\ref{tab:hierarchical_ucm_grouped} reports OA scores for a two-level classification under various clustering and prompting strategies. At the top level ($D=0$), the highest OA is achieved when using GPT-4o to cluster and classify, yielding OA scores of $0.864$ (Kosmos 2) and $0.807$ (Llama-3.2). Supplying the full class list in the prompt still reduces OA scores substantially, in some cases by $\approx10$ percentage points or more. This confirms that long enumeration dilutes the model's semantic focus.  At the fine level ($D=1$), OA falls by roughly $0.20$–$0.30$ across all settings, reflecting the increased difficulty of distinguishing among finer-grained and more similar classes.

The hierarchical design mitigates confusion among highly similar classes by grouping them. As a result, the model achieves stronger performance on the meta-classes (e.g., for $D=0$, with GPT-4o used both for clustering and as classifier, OA scores $\geq 0.757$).  
Finally, meta-classes from GPT-4o achieved greater OA than meta-classes from LLaMA 3.1 across all settings.

\begin{table}[tbp!]
\centering
\caption{Overall Accuracy (OA) for RESISC45 satellite image patches, labeled based on ground truth classes, evaluated in a zero-shot setting and compared with a fine-tuned ResNet50 with ImageNet weights baseline from~\cite{corley2024revisiting}. Best OA scores for each classifier and vision model from GeoVision Labeler (GVL) are highlighted in bold. The \textit{Classes} column indicates whether the list of classes is inserted into the vLLM's prompt. \textit{Other} refers to vision models that are used in a standalone (i.e., not part of GVL).}
\resizebox{0.5\textwidth}{!}{%
\begin{tabular}{lccccc}
\toprule
&& & \multicolumn{3}{c}{\textbf{Vision Model}} \\
\cmidrule(lr){4-6}
{\textbf{Pipeline}} & {\textbf{Classifier}} & {\textbf{Classes}} & \multicolumn{1}{c}{\textbf{Other}} & \multicolumn{1}{c}{\textbf{Kosmos 2}} & \multicolumn{1}{c}{\textbf{Llama 3.2}} \\
\midrule
\multirow{1}{*}{CLIP} & - & - & 0.610 & - & - \\
\cmidrule(lr){1-6}
\multirow{1}{*}{ResNet50 (ImageNet)} & \multirow{2}{*}{-} & \multirow{2}{*}{-} & \multirow{2}{*}{0.775} & \multirow{2}{*}{-} & \multirow{2}{*}{-} \\
\cite{corley2024revisiting} & & & & & \\
\multirow{1}{*}{RS-CLIP~\cite{li2023rs}} & - & - & 0.858 & - & - \\
\cmidrule(lr){1-6}
\multirow{6}{*}{GVL (Ours)} & \multirow{2}{*}{Llama-3.1} & \textcolor{black}{$\checkmark$} & - & 0.272 & \textbf{0.450} \\
& & \textcolor{black}{$\times$} & - & \textbf{0.476} & 0.440 \\
\cmidrule(lr){2-6}
& \multirow{2}{*}{Phi-3} & \textcolor{black}{$\checkmark$} & - & 0.344 & 0.461 \\
& & \textcolor{black}{$\times$} & - & \textbf{0.523} & \textbf{0.473} \\
\cmidrule(lr){2-6}
& \multirow{2}{*}{GPT-4o} & \textcolor{black}{$\checkmark$} & - & 0.352 & \textbf{0.540} \\
& & \textcolor{black}{$\times$} & - & \textbf{0.565} & 0.539 \\
\bottomrule
\end{tabular}%
}
\label{tab:resisc45}
\end{table}

\textbf{RESISC45:}
As presented in Table~\ref{tab:resisc45}, CLIP already reaches $0.610$ in OA--about $80$\% of a fine-tuned ResNet50 and over $70$\% of RS-CLIP ($0.858$)~\cite{corley2024revisiting, li2023rs}. 
The best GVL OA score of 0.565 is achieved by combining the Kosmos 2 vLLM with the GPT-4o classifier, using a prompt that excludes class enumeration (Classes = $\times$).
Indeed, forcing all $45$ classes into the vLLM prompt usually lowers accuracy (e.g., Kosmos 2+Llama-3.1 drops: $0.476 \rightarrow 0.272$; Phi-3: $0.523\rightarrow0.344$), showing similar patterns with UC Merced.

Table~\ref{tab:hierarchical_resisc45_grouped} reports OA scores for three-level hierarchical classification on RESISC45 under different clustering and prompt-engineering strategies. At the top level ($D=0$), the single best performer is clustering and classifying with GPT-4o, reaching $0.843$ (Kosmos 2) and $0.823$ (Llama 3.2). Injecting the full class list into the prompt generally lowers OA, especially for Kosmos 2. For example, the highest Kosmos 2 OA score drops from $0.843$ to $0.809$. The confusion matrix for GVL with Llama 3.2 as vLLM and GPT-4o as classifier, including classes (Classes=$\checkmark$), (see Figure \ref{fig:cm-resisc45-llama32-gpt4o-classes} in the Appendix) suggests that the lower performance of GVL on RESISC45 is also likely due to confusion among visually similar categories within the 45 classes.

\begin{table}[ht!]
\centering
\caption{Overall Accuracy for hierarchical classification of RESISC45 satellite images. The original 45 classes were clustered into 4 meta-classes at depth $D=0$ and 3 meta-classes at depth $D=1$. $D=2$ refers to the original classes. The \textit{Classes} column indicates whether the list of original classes was inserted into the vLLM prompt. Best OA scores for each classifier and vision model are highlighted in bold. The \textit{Classes} column indicates whether the list of classes is inserted into the vLLM's prompt.}
\resizebox{0.5\textwidth}{!}{%
\begin{tabular}{l l c c c c}
\toprule
 & & & & \multicolumn{2}{c}{\textbf{Vision Model}} \\
 \cmidrule(lr){5-6}
\textbf{Clustering} & \textbf{Classifier} & \textbf{Classes} & \textbf{Depth} & \textbf{Kosmos 2} & \textbf{Llama 3.2} \\
\midrule
\multirow{18}{*}{Llama-3.1}
  & \multirow{6}{*}{Llama-3.1}
    & \multirow{3}{*}{$\times$} & 0 & \textbf{0.628} & 0.563 \\
  &    &                             & 1 & 0.486 & 0.423 \\
  &    &                             & 2 & 0.330 & 0.289 \\
  \cmidrule(lr){3-6}
  &    & \multirow{3}{*}{$\checkmark$} & 0 & 0.571 & \textbf{0.576} \\
  &    &                             & 1 & 0.398 & 0.400 \\
  &    &                             & 2 & 0.257 & 0.249 \\
 \cmidrule(lr){2-6}
  & \multirow{6}{*}{Phi-3}
    & \multirow{3}{*}{$\times$} & 0 & \textbf{0.694} & \textbf{0.650} \\
  &    &                             & 1 & 0.561 & 0.501 \\
  &    &                             & 2 & 0.370 & 0.331 \\
  \cmidrule(lr){3-6}
  &    & \multirow{3}{*}{$\checkmark$} & 0 & 0.608 & 0.621 \\
  &    &                             & 1 & 0.461 & 0.442 \\
  &    &                             & 2 & 0.308 & 0.280 \\
 \cmidrule(lr){2-6}
  & \multirow{6}{*}{GPT-4o}
    & \multirow{3}{*}{$\times$} & 0 & \textbf{0.606} & 0.581 \\
  &    &                             & 1 & 0.494 & 0.470 \\
  &    &                             & 2 & 0.325 & 0.323 \\
  \cmidrule(lr){3-6}
  &    & \multirow{3}{*}{$\checkmark$} & 0 & 0.557 & \textbf{0.603} \\
  &    &                             & 1 & 0.421 & 0.465 \\
  &    &                             & 2 & 0.260 & 0.307 \\
\midrule
\multirow{18}{*}{GPT-4o}
  & \multirow{6}{*}{Llama-3.1}
    & \multirow{3}{*}{$\times$} & 0 & \textbf{0.818} & 0.803 \\
  &    &                             & 1 & 0.586 & 0.563 \\
  &    &                             & 2 & 0.390 & 0.371 \\
  \cmidrule(lr){3-6}
  &    & \multirow{3}{*}{$\checkmark$} & 0 & 0.790 & \textbf{0.810} \\
  &    &                             & 1 & 0.550 & 0.577 \\
  &    &                             & 2 & 0.350 & 0.365 \\
 \cmidrule(lr){2-6}
  & \multirow{6}{*}{Phi-3}
    & \multirow{3}{*}{$\times$} & 0 & \textbf{0.828} & \textbf{0.801} \\
  &    &                             & 1 & 0.592 & 0.530 \\
  &    &                             & 2 & 0.401 & 0.346 \\
  \cmidrule(lr){3-6}
  &    & \multirow{3}{*}{$\checkmark$} & 0 & 0.811 & 0.791 \\
  &    &                             & 1 & 0.571 & 0.541 \\
  &    &                             & 2 & 0.358 & 0.341 \\
 \cmidrule(lr){2-6}
  & \multirow{6}{*}{GPT-4o}
    & \multirow{3}{*}{$\times$} & 0 & \textbf{0.843} & 0.823 \\
  &    &                             & 1 & 0.648 & 0.622 \\
  &    &                             & 2 & 0.453 & 0.446 \\
  \cmidrule(lr){3-6}
  &    & \multirow{3}{*}{$\checkmark$} & 0 & 0.809 & \textbf{0.825} \\
  &    &                             & 1 & 0.618 & 0.635 \\
  &    &                             & 2 & 0.413 & 0.441 \\
\bottomrule
\end{tabular}%
}
\label{tab:hierarchical_resisc45_grouped}
\end{table}

The hierarchical paradigm nonetheless alleviates confusion: coarse-level grouping (e.g., $D=0$) yields strong performance (OA up to $0.843$), demonstrating the benefit of decomposing the complex 45-way task into progressively refined stages.
Moving to the mid-level ($D=1$), OA uniformly falls--by as much as $\approx0.20$ in some cases. The best scores around $0.648$ with Kosmos 2 and $0.635$ with Llama 3.2, reflect the added challenge of separating finer meta-groups. At the finest granularity, ($D=2$), OA degrades further (best around $0.453$ with Kosmos 2 and $0.446$ with Llama 3.2), underscoring the difficulty of discriminating among the original 45 classes. Similarly to UC Merced hierarchical classification results, meta-classes from GPT-4o achieved greater OA than meta-classes from LLaMA 3.1 across all settings.  

\section{Discussion}
Across all three datasets--SpaceNet v7, UC Merced, and RESISC45--GVL's two-stage pipeline consistently demonstrates the strength of combining a vLLM model with an LLM classifier. Common to every dataset is the sensitivity to prompt design: enumerating target classes in the prompt can improve performance when the label set is small (e.g., binary building detection). But accuracy lowers as the number of classes grows (e.g., 21 classes in UC Merced, 45 in RESISC45). Moreover, injecting geo-context can boost performance in the binary SpaceNet v7 classification task, when classes are also provided--e.g., raising Kosmos 2's OA from 0.859 to 0.910 and GPT-4o's from 0.878 to 0.917--though such metadata may not always be available.

For complex multi-class scenarios, the hierarchical classification strategy provides a unified way to mitigate confusion among visually similar labels. At the coarsest level ($D=0$), GVL attains its highest overall accuracies (up to 0.864 on UC Merced, and 0.843 on RESISC45), showing that grouping semantically related classes into a smaller set of meta-classes simplifies the classification task. Performance then tapers at finer levels ($D=1,2$), with OA scores consistently dropping across all settings, reflecting the increasing difficulty of distinguishing among more granular categories.

Furthermore, the class clustering method that employed GPT-4o consistently achieved higher OA scores than the one based on LLaMA 3.1 across all evaluation settings. This highlights how crucial the choice of class clustering method is to the effectiveness of the hierarchical classification strategy.

These results illustrate that GVL's two-stage architecture, light prompt engineering, and optional hierarchical grouping form a versatile toolkit for zero-shot remote sensing classification. It scales from simple binary tasks to large, fine-grained taxonomies where hierarchical classification provides a competitive baseline for weak label generation.

\section{Limitations}

Despite these promising findings, GVL's zero-shot nature and current implementation impose several constraints. One is prompt sensitivity: while brief, targeted prompts can sharpen the model's focus, long class enumerations--especially with dozens of categories--readily dilute the vLLM's descriptive capacity, necessitating dataset-specific prompt tuning. Hierarchical classification reduces confusion from similar labels at the meta-classes levels but introduces extra inference stages and complexity, which may incur latency, increase computational cost, and complicate deployment pipelines. Moreover, because GVL operates without task-specific fine-tuning, it cannot fully match the performance of supervised or further tuned models.

Model dependency further limits robustness as the quality of generated descriptions and classifications hinges on the selected vision and language models. Suboptimal or domain-misaligned models may produce inaccurate or misleading outputs. Likewise, image characteristics play a critical role. Small image patches (e.g., below $224\times224$ pixels) can lack sufficient context for accurate description, while very large patches may overwhelm the vLLM's attention and dilute focus on discriminative details. Very high-resolution images may require tiling into smaller chunks, potentially fragmenting spatial context and slowing down the pipeline. Additionally,  vLLMs are typically not trained on remote sensing data, so unusual environmental conditions--cloud cover, seasonal vegetation changes, or sensor artifacts--can degrade performance. Finally, GVL currently supports only RGB inputs, yet many satellite imagery platforms provide additional spectral bands (e.g., near-infrared) that could enhance classification. Leveraging vLLMs designed for multi-spectral data represents a promising direction to overcome this format restriction and further improve zero-shot remote sensing performance.  

\section{Conclusion}
As satellite imagery becomes more accessible, the need for effective classification and interpretation--particularly in low-resource environments--has become increasingly pressing.
GVL addresses this gap by leveraging the synergistic capabilities of vLLMs and LLMs. By generating and classifying textual descriptions, the framework offers a flexible and adaptable solution for classifying satellite imagery without the need for task-specific training data. It demonstrates generalizability--particularly in low- and separable-class scenarios--achieving up to 93.2 \% OA on SpaceNet v7 when distinguishing between \emph{Buildings} and \emph{No Buildings} image patches in a purely zero-shot setting.

A key extension of GVL is its hierarchical classification strategy for complex multi-class datasets (UC Merced, RESISC45). By clustering original labels into a small number of meta-classes, GVL attains coarse-level OA scores as high as $0.864$ on UC Merced and $0.843$ on RESISC45 datasets, before degrading at finer depths ($D=1,2$). This hierarchical decomposition mitigates confusion among visually similar classes and yields more interpretable error patterns.

The modular GVL pipeline supports plug-and-play combinations of vision and language models, delivering true zero-shot classification in low-resource environments, with decisions traceable to the vLLM's generated descriptions. Nonetheless, challenges remain: GVL's performance depends on high-quality vLLMs and LLMs; small image patches (e.g., below $224\times224$ pixels) may lose critical context; and the current RGB-only design cannot exploit non-visible spectral bands common in satellite imagery. Future work will explore integrating vLLMs pretrained on multi-spectral remote sensing data and adaptation strategies to further close the gap to supervised baselines.
GVL is open-sourced on GitHub\footnote{\url{https://github.com/microsoft/geo-vision-labeler}} to foster community contributions.

\bibliography{biblio}
\bibliographystyle{icml2025}

\newpage
\appendix
\onecolumn
\counterwithin{figure}{section} 
\counterwithin{table}{section} 

\section{Appendix}
\begin{figure}[ht!]
    \centering
    \includegraphics[width=1\linewidth]{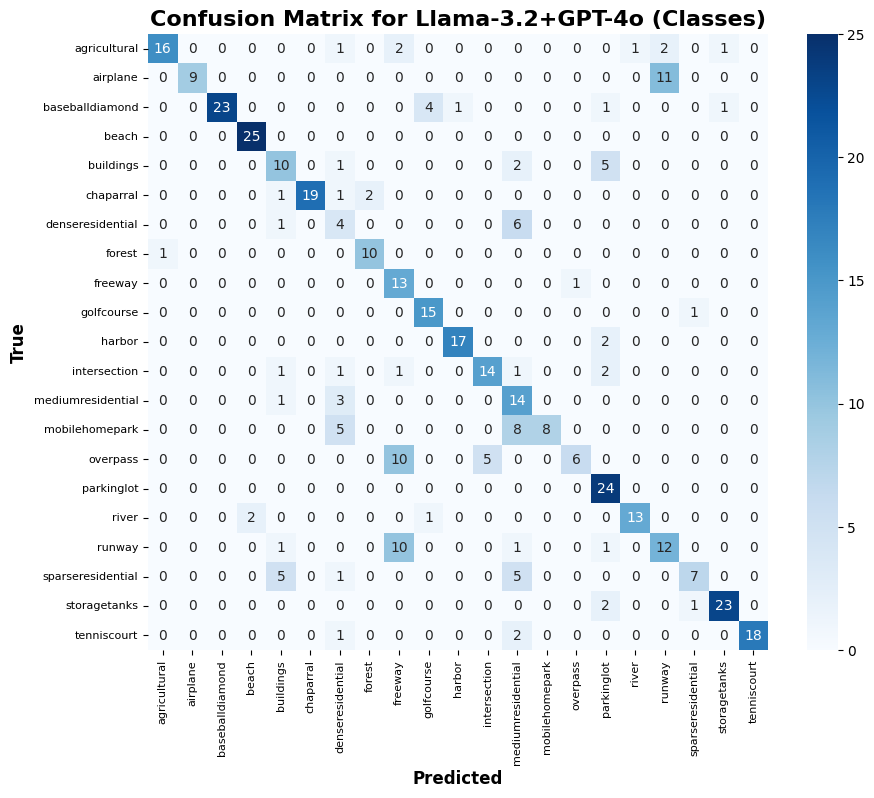}
    \caption{Confusion matrix for the UC Merced classification task with the GeoVision Labeler (GVL) pipeline with Llama 3.2 as vLLM and GPT-4o as classifier. \textit{Classes} are included in the vLLM's prompt.}
    \label{fig:cm-ucm-llama32-gpt4o-classes}
\end{figure}

\begin{figure}[ht!]
    \centering
    \includegraphics[width=1\linewidth]{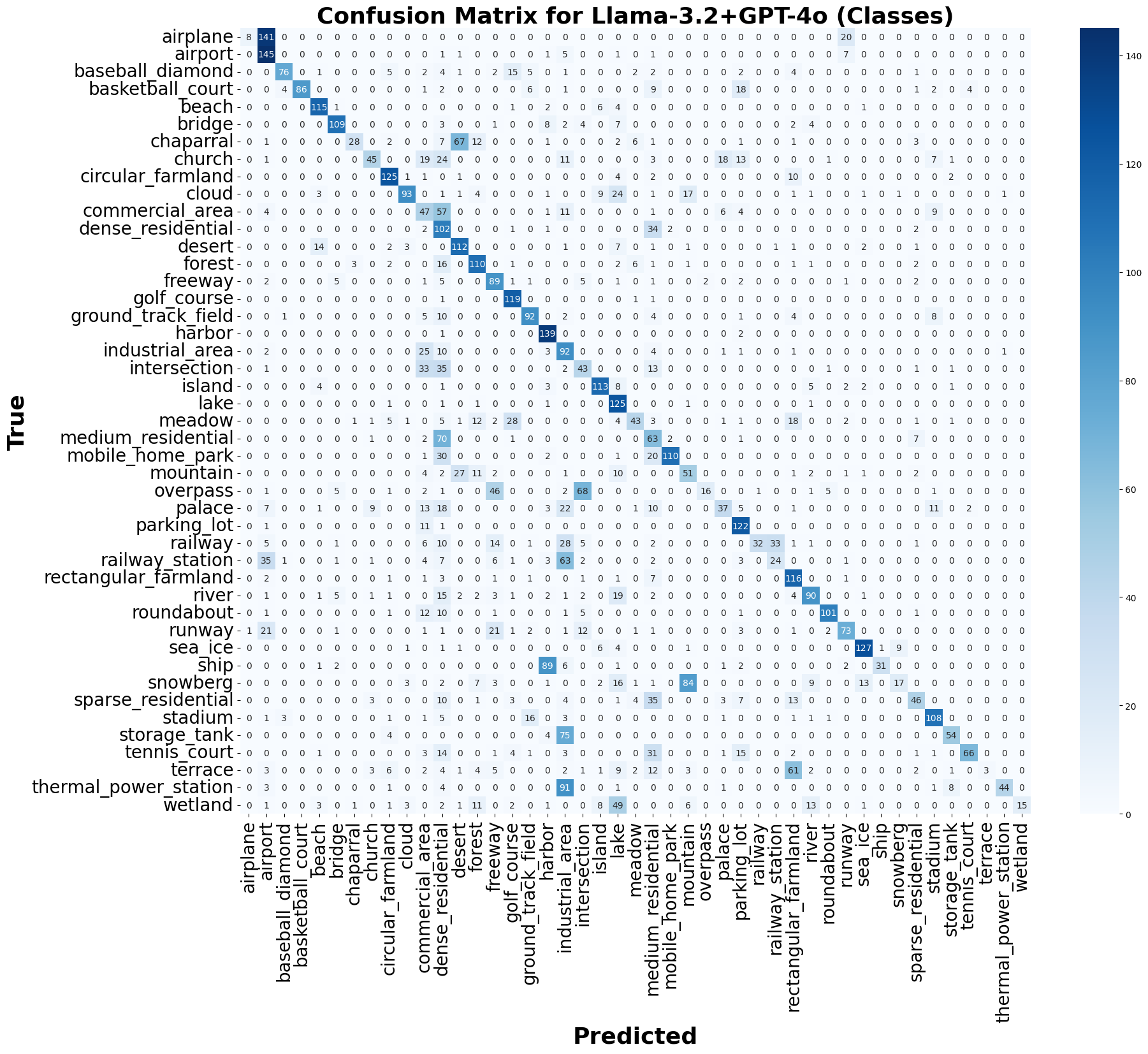}
    \caption{Confusion matrix for the RESISC45 classification task with the GeoVision Labeler (GVL) pipeline with Llama 3.2 as vLLM and GPT-4o as classifier. \textit{Classes} are included in the vLLM's prompt.}
    \label{fig:cm-resisc45-llama32-gpt4o-classes}
\end{figure}

\begin{table}[ht!]
\centering
\caption{Meta-classes derived from the 21 original UC Merced classes at depths $D=0$ (meta-class) and $D=1$ (original classes) for two clustering models.}
\resizebox{\textwidth}{!}{%
\begin{tabular}{l l p{8cm}}
\toprule
\textbf{Model} & \textbf{Meta-class ($D=0$)} & \textbf{Classes ($D=1$)} \\
\midrule
\multirow{5}{*}{Llama-3.1}
  & Manmade Structures        & \texttt{baseballdiamond}, \texttt{buildings}, \texttt{harbor}, \texttt{intersection}, \texttt{overpass}, \texttt{parkinglot}, \texttt{storagetanks} \\
\cmidrule(lr){2-3}
  & Natural Environments      & \texttt{agricultural}, \texttt{beach}, \texttt{chaparral}, \texttt{forest}, \texttt{river} \\
\cmidrule(lr){2-3}
  & Recreational Facilities   & \texttt{golfcourse}, \texttt{tenniscourt} \\
\cmidrule(lr){2-3}
  & Residential Areas         & \texttt{denseresidential}, \texttt{mediumresidential}, \texttt{mobilehomepark}, \texttt{sparseresidential} \\
\cmidrule(lr){2-3}
  & Transportation Infrastructure  
                             & \texttt{airplane}, \texttt{freeway}, \texttt{runway} \\
\midrule
\multirow{5}{*}{GPT-4o}
  & Natural Landscapes        & \texttt{agricultural}, \texttt{beach}, \texttt{chaparral}, \texttt{forest}, \texttt{river} \\
\cmidrule(lr){2-3}
  & Recreational Areas        & \texttt{baseballdiamond}, \texttt{golfcourse}, \texttt{tenniscourt} \\
\cmidrule(lr){2-3}
  & Residential Areas         & \texttt{denseresidential}, \texttt{mediumresidential}, \texttt{mobilehomepark}, \texttt{sparseresidential} \\
\cmidrule(lr){2-3}
  & Transportation             & \texttt{airplane}, \texttt{freeway}, \texttt{harbor}, \texttt{intersection}, \texttt{overpass}, \texttt{parkinglot}, \texttt{runway} \\
\cmidrule(lr){2-3}
  & Urban Infrastructure      & \texttt{buildings}, \texttt{storagetanks} \\
\bottomrule
\end{tabular}%
}
\label{tab:ucmerced_meta_classes}
\end{table}

\begin{table}[ht!]
\centering
\caption{Meta-classes derived from the 45 original RESISC45 classes at depths $D=0,1$ for two clustering models. $D=2$ refers to the original classes in the dataset.}
\resizebox{\textwidth}{!}{%
\begin{tabular}{l l l p{8cm}}
\toprule
\textbf{Model} & \textbf{Meta-class ($D=0$)} & \textbf{Meta-class ($D=1$)} & \textbf{Classes ($D=2$)} \\
\midrule
\multirow{12}{*}{Llama-3.1}
  & \multirow{3}{*}{Bodies of Water}
    & Aquatic Ecosystems           & \texttt{wetland} \\\cmidrule(lr){3-4}
  &                                & Bodies of Water              & \texttt{lake}, \texttt{river} \\
  \cmidrule(lr){3-4}
  &                                & Glaciers and Ice             & \texttt{sea\_ice}, \texttt{snowberg} \\
\cmidrule(lr){2-4}
  & \multirow{6}{*}{Manmade Structures}
    & Manmade Structures           & \texttt{baseball\_diamond}, \texttt{church}, \texttt{commercial\_area}, \\ 
  &                                &                              & \texttt{dense\_residential}, \texttt{harbor}, \texttt{intersection}, \\ 
  &                                &                              & \texttt{medium\_residential}, \texttt{palace}, \texttt{parking\_lot}, \\ 
  &                                &                              & \texttt{railway\_station}, \texttt{roundabout}, \texttt{ship}, \\ 
  &                                &                              & \texttt{sparse\_residential}, \texttt{stadium}, \texttt{storage\_tank}, \\ 
  &                                &                              & \texttt{terrace}, \texttt{thermal\_power\_station} \\
  \cmidrule(lr){3-4}
  &                                & Natural \& Agricultural Landscapes & \texttt{circular\_farmland}, \texttt{rectangular\_farmland} \\
  \cmidrule(lr){3-4}
  &                                & Transportation Infrastructure & \texttt{airplane}, \texttt{airport}, \texttt{bridge}, \texttt{freeway},\\ 
  &                                &                              & \texttt{overpass}, \texttt{railway}, \texttt{runway} \\
\cmidrule(lr){2-4}
  & \multirow{3}{*}{Natural Landscapes}
    & Coastal \& Isolated Environments  & \texttt{beach}, \texttt{island} \\
    \cmidrule(lr){3-4}
  &                                & Landscapes with Vegetation       & \texttt{chaparral}, \texttt{forest}, \texttt{meadow} \\
  \cmidrule(lr){3-4}
  &                                & Natural Landforms                & \texttt{cloud}, \texttt{desert}, \texttt{mountain} \\
\cmidrule(lr){2-4}
  & \multirow{3}{*}{Recreational \& Industrial Areas}
    & Recreational \& Industrial Areas  & \texttt{industrial\_area} \\
    \cmidrule(lr){3-4}
  &                                & Residential Areas                & \texttt{mobile\_home\_park} \\
  \cmidrule(lr){3-4}
  &                                & Sports Facilities                & \texttt{basketball\_court}, \texttt{golf\_course}, \texttt{ground\_track\_field}, \texttt{tennis\_court} \\
\midrule
\multirow{11}{*}{GPT-4o}
  & \multirow{3}{*}{Natural Landscapes}
    & Agricultural and Terrain          & \texttt{circular\_farmland}, \texttt{rectangular\_farmland} \\
    \cmidrule(lr){3-4}
  &                                & Aquatic and Water Bodies           & \texttt{lake}, \texttt{river}, \texttt{sea\_ice}, \texttt{wetland} \\\cmidrule(lr){2-4}
  &                                & Natural Landscape and Vegetation   & \texttt{beach}, \texttt{chaparral}, \texttt{cloud}, \texttt{desert}, \\ 
  &                                &                                   & \texttt{forest}, \texttt{island}, \texttt{meadow}, \texttt{mountain}, \texttt{snowberg} \\
\cmidrule(lr){2-4}
  & \multirow{3}{*}{Recreational Facilities}
    & Ball Sports                       & \texttt{baseball\_diamond}, \texttt{basketball\_court}, \texttt{stadium}, \\ 
  &                                &                                   & \texttt{tennis\_court} \\
  \cmidrule(lr){3-4}
  &                                & Golf                              & \texttt{golf\_course} \\
  \cmidrule(lr){3-4}
  &                                & Track and Field                   & \texttt{ground\_track\_field} \\
\cmidrule(lr){2-4}
  & \multirow{3}{*}{Transportation \& Infrastructure}
    & Industrial Facilities              & \texttt{ship}, \texttt{storage\_tank}, \texttt{thermal\_power\_station} \\
    \cmidrule(lr){3-4}
  &                                & Traffic Structures                & \texttt{bridge}, \texttt{intersection}, \texttt{overpass}, \\ 
  &                                &                                   & \texttt{roundabout} \\
  \cmidrule(lr){3-4}
  &                                & Transportation Infrastructure      & \texttt{airplane}, \texttt{airport}, \texttt{freeway}, \texttt{harbor}, \\ 
  &                                &                                   & \texttt{parking\_lot}, \texttt{railway}, \texttt{railway\_station}, \\ 
  &                                &                                   & \texttt{runway} \\
\cmidrule(lr){2-4}
  & \multirow{2}{*}{Urban Structures}
    & Commercial \& Industrial Zones     & \texttt{commercial\_area}, \texttt{industrial\_area} \\
    \cmidrule(lr){3-4}
  &                                & Public \& Historic Buildings       & \texttt{church}, \texttt{palace} \\
  \cmidrule(lr){3-4}
  &                                & Residential Areas                  & \texttt{dense\_residential}, \texttt{medium\_residential}, \\ 
  &                                &                                   & \texttt{mobile\_home\_park}, \texttt{sparse\_residential}, \texttt{terrace} \\
\bottomrule
\end{tabular}%
}
\label{tab:resisc45_meta_classes}
\end{table}


\end{document}